\documentclass{article}

\usepackage{microtype}
\usepackage{graphicx}
\usepackage{subfigure}
\usepackage{booktabs} 
\usepackage[most]{tcolorbox}
\usepackage{float}
\usepackage{placeins}
\usepackage{graphicx}
\usepackage{caption}
\usepackage{subcaption}

\usepackage{multicol}

\usepackage{hyperref}

\usepackage[accepted]{icml2025} 

\usepackage{amsmath}
\usepackage{amssymb}
\usepackage{mathtools}
\usepackage{amsthm}

\usepackage[capitalize,noabbrev]{cleveref}

\theoremstyle{plain}

\theoremstyle{definition}

\theoremstyle{remark}

\usepackage[textsize=tiny]{todonotes}
\usepackage{listings}         

\begin{document}
\twocolumn[
\icmltitle{Rewrite-to-Rank: Optimizing Ad Visibility via Retrieval-Aware Text Rewriting}



\icmlsetsymbol{equal}{*}

\begin{icmlauthorlist}
\icmlauthor{Chloe Ho}{}
\icmlauthor{Ishneet Sukhvinder Singh}{sch}
\icmlauthor{Diya Sharma}{equal}
\icmlauthor{Tanvi Reddy Anumandla}{equal}
\icmlauthor{Michael Lu}{sch3}
\icmlauthor{Vasu Sharma}{sch4}
\icmlauthor{Kevin Zhu}{comp}
\end{icmlauthorlist}

\icmlaffiliation{sch}{University of Oxford, Oxford, England}
\icmlaffiliation{sch3}{University of Berkeley, Berkeley, CA, USA}
\icmlaffiliation{sch4}{Carnegie Mellon University, Pittsburgh, PA, USA}
\icmlaffiliation{comp}{Algoverse AI Research, Remote}

\icmlcorrespondingauthor{Chloe Ho}{chloe.ho.0830@gmail.com}

\icmlkeywords{Retrieval-Augmented Generation, Advertisement Rewriting, Proximal Policy Optimization, Reinforcement Learning, Large Language Models, Content Ranking, Advertisement Ranking, Content Optimization}

\vskip 0.3in
]



\printAffiliationsAndNotice{\icmlEqualContribution} 
\begin{abstract}
Search algorithms and user query relevance have given LLMs the ability to return relevant information, but the effect of content phrasing on ad visibility remains underexplored. We investigate how LLM-based rewriting of advertisements can improve their ranking in retrieval systems and inclusion in generated LLM responses, without modifying the retrieval model itself. We introduce a supervised fine-tuning framework with a custom loss balancing semantic relevance and content fidelity. To evaluate effectiveness, we propose two metrics: $\Delta$MRR@K (ranking improvement) and $\Delta$DIR@K (inclusion frequency improvement). Our approach presents a scalable method to optimize ad phrasing, enhancing visibility in retrieval-based LLM workflows. Experiments across both instruction-based and few-shot prompting demonstrate that PPO trained models outperform both prompt engineering and supervised fine-tuning in most cases, achieving up to a 2.79 $\Delta$DIR@5 and 0.0073 $\Delta$MRR@5 in instruction-based prompting. These results highlight the importance of how the ad is written before retrieval and prompt format and reinforcement learning in effective ad rewriting for LLM integrated retrieval systems. 
\end{abstract}

\section{Introduction}
\label{introduction}
Large language models (LLMs) have demonstrated impressive capabilities across a wide range of natural language tasks such as question answering and summarization \cite{Nijkamp23,Lewis21}. To overcome the limitations of internal knowledge, many modern LLM systems are designed to use Retrieval-Augmented Generation (RAG) systems \cite{Lewis21}, where a retriever selects relevant documents from an external corpus which is then passed to a generator. As LLMs continue to be used in industries like digital advertising, whose global revenue reached USD 259 billion in 2024, up 15\% from 2023 \cite{IAB25}, their ability to retrieve external documents is often limited by the quality of current retrieval systems \cite{Alinejad24}. Current RAG pipelines focus on query-side optimization, often relying on black-box retrievers that compute similarity scores between user queries and a static document corpus to identify the most relevant documents \cite{Liu24}, yet they ignore the producer side, i.e. how documents themselves can be optimized for retrieval. This oversight leads to relevant advertisements being missed due to minor differences in phrasing, formatting, or keyword choice \cite{Dash24}. Suboptimal ranking in the retrieval process can significantly reduce ad visibility, which hurts revenue potential. Optimizing to fill the producer-side gap via integrating advertisements directly into LLM retrieval workflows presents a promising solution for relevant sponsored content and sustainable revenue models for AI applications.

In this paper, we introduce a novel approach to improving the ranking and inclusion of advertisements by rewriting ad content rather than modifying the retrieval system itself. Specifically, we propose using a fine-tuned language model with a custom-designed loss function to encourage rewritten advertisements to: (1) align rewrites with user queries and (2) preserve the ad's original meaning. The contributions of this paper are as follows:

\begin{itemize}
 \item We propose a new approach that improves document ranking and likelihood of inclusion in LLM output by rewriting ad content, rather than modifying the retrieval system, introducing a complementary angle to existing re-ranking and retriever-based methods.
 \item We develop a tailored loss function that balances maximizing relevance to user queries with preservation of the original intent of the ad, enabling effective fine-tuning for retrieval-sensitive content rewriting.
 \item We show how Proximal Policy Optimization (PPO) can be applied to train LLMs to optimize ad visibility, demonstrating a new use case for RL in language model fine-tuning.
\item We contribute an evaluation setup using $\Delta$MRR@K and $\Delta$DIR@K and highlight a scalable, potentially monetizable application of LLMs in ad retrieval, offering a framework others can adopt for commercial or academic use.
\end{itemize}

\section{Related Works}

Our work builds upon recent progress in Retrieval-Augmented Generation (RAG) systems and the broader literature on optimizing language model outputs for downstream utility.

\textbf{Document Expansion vs. Ad Phrasing.} Traditional document expansion techniques, such as term importance prediction \citep{MacAvaney20} and BERT-based re-ranking \citep{Nogueira20}, often assume neutral objectives that can be generalized to a broad document corpus via optimizing for query relevance. Ad phrasing, conversely, holds even subtle linguistic changes to a higher standard insofar as influencing user engagement. Rather than uniformly enriching content, we target strategic micro-edits that align with the retrieval objectives of black-box models. 

\textbf{Retrieval-Augmented Generation.} RAG has emerged as a powerful framework for enhancing language models with external knowledge sources \citep{Lewis21}, where a retriever selects documents passed to a generator. While prior work has largely focused on improving retrievers \citep{Liu24} or combining retriever-generator feedback loops \citep{Alinejad24}, our approach complements these efforts by rewriting the documents themselves to enhance retrievability—without modifying the retriever.

\textbf{Ad Content Optimization.} In the context of digital advertising, existing work such as \citep{Dash24, article} has highlighted the limitations of sponsored content ranking, showing that minor changes in phrasing can drastically affect ad visibility. However, most methods address this through ranking adjustments or changes on the platform side. In contrast, we propose content-side optimization via LLM-driven rewriting, allowing advertisers to improve discoverability in black-box systems.

\textbf{Prompting and Language Model Steering.} Our use of Chain-of-Thought prompting \citep{Wei23} and few-shot learning techniques \citep{Brown20} follows recent trends in LLM control. While prior work focuses on task accuracy or reasoning depth, we apply these techniques to guide ad rewriting for improved ranking and inclusion in LLM responses.

\textbf{Reinforcement Learning with LLMs.} We introduce a novel application of Proximal Policy Optimization (PPO) to optimize ad visibility based on a composite reward that balances ranking relevance and content fidelity. Although PPO has been successfully applied to instruction tuning and retrieval optimization \citep{ouyang2022traininglanguagemodelsfollow, kulkarni2024reinforcementlearningoptimizingrag}, its application to retrieval-centric rewriting remains underexplored.

Our framework fills a gap between search ranking systems and LLM generation, providing a scalable method for advertisers to increase content discoverability in AI-driven platforms.

\section{Methods}
Our objective is to rewrite advertisement documents such that they rank higher in retrieval systems and are more likely to be included in the LLM's response, while preserving the semantic content of the original ad.

In order to train the model for document rewriting, we employ a custom loss function, which 1) maximizes the similarity between the rewritten document and the query, 2) ensures that the rewritten document is more similar to the query than randomly selected top k documents, and 3) the rewritten document remains semantically faithful to the original document. 

To interpret the overall performance of the model for each training iteration, we implemented two evaluation metrics MRR@K Improvement ($\Delta$MRR@K) and document inclusion rate @ K improvement ($\Delta$DIR@K), which measure the change in rank a document experiences after a rewriting strategy is applied and the change in frequency in which the document is included in the LLM output, respectively. 

To access inclusion, we prompt for a black-box LLM, Gemini 1.5 Pro, with the original and rewritten ad documents. The prompt is provided in the Appendix and visible in the code base and is written to be intentionally minimal to avoid prompt-induced bias and reflect practical scenarios. 

\subsection{Data Construction}
First, we randomly sampled 11,000 advertisements from the original dataset, CommercialAdsDataset by Microsoft, with 10,000 ads for the train data and 1,000 ads as our test data to construct our initial advertisement dataset. 

Each ad is associated with a domain and subdomain label using a large language model (Gemini 1.5 Pro), which were verified through human annotators with 85\% accuracy. We create a query data set by generating ten queries for each pair of domain-subdomains. For each query, we retrieve the top k relevant ads from the main advertisement dataset and generate LLM responses conditioned on those ads. The retrieval rankings produced for each query are then retained for later evaluation in rewriting strategies.

\subsection{Retriever Configuration}

\textbf{Embedding model.}  
We embed every advertisement \emph{and} every query with the public
Sentence-Transformers checkpoint
\texttt{all-MiniLM-L6-v2}\footnote{384-dimensional output; sentence-transformers v2.6.1.}
Vectors are $\ell_{2}$-normalised so cosine similarity reduces to
inner-product distance.

\textbf{Index type.}  
All 11\,000 ad embeddings (10 k train + 1 k test) are stored in a
\texttt{faiss::IndexFlatIP} (exact inner-product search) built with
FAISS.  
Exact search avoids ANN hyper-parameters and guarantees deterministic
recall. All 11\,000 ads are embedded in mini-batches of 64 and then added to the
index in one call (\verb|index.add(embs)|).  At search time we encode one
query at a time, so the FAISS call sees a batch of 1.

\textbf{LLM for inclusion.}  
Document-inclusion is evaluated by
\texttt{gemini-1.5-pro-001} (Gemini 1.5 Pro, February 2024 release);
the exact prompt appears in Appendix A.

\subsection{Loss Function Construction and Implementation}
The loss function guides the model during training by quantifying how well a rewritten ad balances three goals in a 1:1:1 ratio:
\begin{itemize}
\item Maximize similarity to the user query
\item Maximize difference from top-k documents
\item Preserve semantic meaning from the original ad
\end{itemize}

Let the loss function be composed of the three components:
\begin{itemize}
\item $L_{\text{rel\_gain}}$: Relevance gain loss
\item $L_{\text{triplet}}$: Triplet sampling loss
\item $L_{\text{fidelity}}$: Content fidelity loss
\end{itemize}

By minimizing loss, the model learns to generate rewrites that aim to improve retrievability while staying loyal to the original content. The overall loss function is a summation:

$L_{\text{total}}$ = $L_{\text{rel\_gain}}$ + $L_{\text{triplet}}$ + $L_{\text{fidelity}}$

Let:

\( d \) = the document we are tracking

\( Q_d \) = set of user queries relevant to \( d \)

\subsubsection{Relevance Gain Loss ($L_{\text{rel\_gain}}$)}
This term encourages the rewritten ad to be more relevant to the query than the original version. Let:

$d_{\text{before}}$: original advertisement

$d_{\text{after}}$: rewritten advertisement

$q \in Q_{\text{d}}:$ a query relevant to the ad $d$

sim(a,b): cosine similarity between the embeddings of $a$ and $b$

Then: 

$L_{\text{rel\_gain}}(q,d)$ = $-$(sim($q,d_{\text{after}}$) - sim($q,d_{\text{before}}$))

\subsubsection{Triplet Sampling Loss ($L_{\text{triplet}}$)}
This loss term encourages the rewritten ad to be more relevant to the query than competing documents retrieved for the same query. We sample a set of 3 documents from the top-k retrieved list for a given query and compute their average similarity to the query. This average is then compared with the similarity between the rewritten ad and the query. This loss penalizes rewritten documents whose relevance to the query falls below the average of a sample of the top-k retrieved documents.

$N_{q,d} : \text{set of 3} \text{ documents for query } q$

Define the average similarity of the distractors to the query as:

$s_{\lnot}(q) = \frac{1}{3} \sum_{n \in N_{q,d}} sim(n, q)$

Then the triple sampling loss is:

$L_\text{triplet}(q, d) = -\left[ sim(q, d_{\text{after}}) - sim(q, s_{\lnot}(q)) \right]$

\subsubsection{Content Fidelity ($L_{\text{fidelity}}$)}
We preserve content fidelity between the rewritten and original documents, maximizing the similarity between the original and rewritten documents. 

$L_\text{fidelity}(q, d) = 1 - sim(d_{\text{after}}, d_{\text{before}})$

Our overall loss function for each document is as follows:

$L_{\text{total}}$ = $L_{\text{rel\_gain}}$ + $L_{\text{triplet}}$ + $L_{\text{fidelity}}$

\subsection{Prompting Strategies}
To guide the language model in rewriting advertisements effectively, we designed two prompts based on a technique called Chain-of-Thought (CoT) reasoning. This approach encourages the model to "think out loud" by walking through the reasoning process before generating a final rewrite. 

Using two prompting strategies grounded in popular, established prompting methods \cite{Wei23,Brown20}, our goal was to help the model focus on improving how relevant an ad sounds to a user’s query, without losing the original meaning. The specific prompts are provided in the Appendix.

\subsubsection{Strategy A: Instruction-based Prompting with CoT (Zero-Shot):}

In this approach, the model is given a high-level instruction to rewrite an ad in a way that improves its retrieval ranking and likelihood of inclusion and is told to reason and think before responding. The prompt encourages semantic alignment with user intent and explicit reasoning before rewriting.

\subsubsection{Strategy B: Few-Shot Prompting with CoT:}
This setup provides the LLM with two examples of ad rewrites, including the original ad, rationale, and improved version. After observing the examples, the model is prompted to perform a similar rewrite on a new ad input.

\section{Experiments}
\subsection{Evaluation Metrics}

Let: 

\( K \) = rank cut-off we impose (e.g. 10)

\( v \in \{\text{before}, \text{after}\} \rightarrow \) document version

\(\text{rank}_v(d, q)\) = rank of \(d\) for query \(q\)

\( X_v^{(d,q)} = 1 \) if the LLM actually uses \(d\) in its final answer, else 0

\subsubsection{Metric 1: MMR@K Improvement ($\Delta$MRR@K)}
We evaluate whether a rewriting strategy is effective in improving a document’s ranking by measuring the change in mean reciprocal ranking prior and post imposition. We compute:

$\Delta \text{MRR}@K = \text{MRR}@K_{\text{after}} - \text{MRR}@K_{\text{before}}$
\[
\Delta \text{MRR}@K = \frac{1}{|Q_d|} \sum_{q \in Q_d} \left( RR_K^{\text{after}}(d, q) - RR_K^{\text{before}}(d, q) \right)
\]
where
\[
RR_K^v(d, q) =
\begin{cases}
\frac{1}{\text{rank}_v(d, q)}, & \text{if } \text{rank}_v(d, q) \leq K \\
0, & \text{otherwise}
\end{cases}
\]

\subsubsection{Metric 2: Document-Inclusion Rate @ K Improvement ($\Delta$DIR@K)}

We track the inclusion rate of a document in the LLM’s response across relevant queries and compute the change between the inclusion rate of a document remaining in the top-K set before and after rewriting. 

This metric represents the absolute change (percentage-point lift) and formally, we define it as follows:

\[
\Delta DIR@K = DIR@K_{\text{after}} - DIR@K_{\text{before}}
\]

Below is a formal definition of the base metric, \( DIR@K_v \):

First, restrict the query set to include only queries where the document rank is within the top-k cut-off in both versions:
\[
Q_d^{(K)} = \left\{ q \in Q_d \;\middle|\; \text{rank}_{\text{before}}(d, q) \leq K \land \text{rank}_{\text{after}}(d, q) \leq K \right\}
\]

Then the Document-Inclusion Rate @ K for version \( v \) of any document is:

\[
DIR@K_{v \in \{\text{before}, \text{after}\}} = \frac{1}{|Q_d^{(K)}|} \sum_{q \in Q_d^{(K)}} X_v^{(d,q)}
\]

\subsection{Baselines}
\subsubsection{Zero-Rewrite Control}
We establish a baseline using the original, unmodified ads. For each query, we evaluate the responses retrieved and generated using the original ad set, without any LLM rewriting.  

\subsubsection{Prompt Engineering}
Each of the two prompting strategies in Section 3.3 are used to generate rewrites for the 1,000 test ads. For each rewritten version, we compute $\Delta$MRR@K and $\Delta$DIR@K relative to the original retrieval and inclusion statistics for each ad.

\subsubsection{Supervised Fine-Tuning (SFT)}
We fine-tune a pretrained language model (LLaMA-3.1-8B) using the original ads as inputs with rewritten advertisements as the ground truths. The rewritten versions are generated through a general prompt, which is provided in the Appendix, that is known to boost metric scores. 

The objective is to teach the model to rewrite an original ad similarly to an improved version which is more likely to rank higher in retrieval and be included in LLM output.

\subsection{Proximal Policy Optimization (PPO)}
To further enhance the retrievability of rewritten ads, we apply Proximal Policy Optimization (PPO) on top of our supervised fine-tuned LLaMA-3.1-8B model. PPO enables the model to optimize directly for our task-specific objective by learning from a custom reward signal.

The reward is defined as the negative of the total loss introduced in Section 3.2. This loss function balances three priorities: improving query relevance, outperforming distractor documents, and preserving content fidelity.

For each ad, we perform the following steps to compute the reward:
\begin{itemize}
\item Identify up to three queries that share the same domain and subdomain as the ad.
\item For each query, retrieve the top-k documents from the original dataset.
\item Compute the reward using the loss function defined in Section 3.2.
\item Average the loss across all selected queries to obtain the final reward signal.
\end{itemize}

This reinforcement learning stage allows the model to go beyond pattern imitation and optimize rewrites that are more likely to rank higher in retrieval and be included in downstream LLM outputs.

\noindent
We implement LoRA fine-tuning with rank~8 on \texttt{all} attention and
MLP projections, enable gradient checkpointing, train in \texttt{bfloat16},
and employ DeepSpeed ZeRO‐2 with CPU optimizer offloading
(\texttt{zero\_stage=2}). All PPO hyper-parameters can be found in the Appendix.

\section{Results}
\subsection{Comparing the Models}
Tables \ref{tab:mrr-dir-original}, \ref{tab:mrr-dir-valuesinst}, and \ref{tab:mrr-dir-valuesfew} present the $\Delta$MRR@5 and $\Delta$DIR@5 for the original ad baseline, prompt engineering, supervised fine-tuning, and PPO training across both prompting strategies, instruction-based and few-shot, for the latter three models. 

\begin{table}[H]
\centering
\begin{tabular}{lcc}
\toprule
\textbf{Model} & \textbf{$\Delta$MRR@5} & \textbf{$\Delta$DIR@5} \\
\midrule
Zero-Rewrite & 0.00 & -0.0807 \\
\bottomrule
\end{tabular}
\caption{Values of $\Delta$MRR and $\Delta$DIR for the zero-rewrite baseline for k = 5}
\label{tab:mrr-dir-original}
\end{table}

\vspace{1em}
\begin{table}[H]
\centering
\begin{tabular}{lcc}
\toprule
\textbf{Model} & \textbf{$\Delta$MRR@5} & \textbf{$\Delta$DIR@5} \\
\midrule
Prompt Eng. & 0.0051 & 0.9061 \\
SFT & 0.0022 & 1.6382 \\
PPO & 0.0073 & 2.7944 \\
\bottomrule
\end{tabular}
\caption{Comparison of $\Delta$MRR and $\Delta$DIR values across different models for k = 5 for instruction-based prompting}
\label{tab:mrr-dir-valuesinst}
\end{table}
\vspace{-1em}
\begin{table}[H]
\centering
\begin{tabular}{lcc}
\toprule
\textbf{Model} & \textbf{$\Delta$MRR@5} & \textbf{$\Delta$DIR@5} \\
\midrule
Prompt Eng. & 0.0067 & 1.9133 \\
SFT & -0.0017 & 1.4199 \\
PPO & -0.0008 & 1.9267 \\
\bottomrule
\end{tabular}
\caption{Comparison of $\Delta$MRR and $\Delta$DIR values across different models for k = 5 for few-shot prompting}
\label{tab:mrr-dir-valuesfew}
\vspace{-1.5em}
\end{table}

In our evaluation, $\Delta$MRR@5 is 0 and $\Delta$DIR@5 is -0.0807 for the zero-rewrite baseline. As expected, both are closer to zero than the metrics when the ads are rewritten. 

Under instruction-based prompting, PPO consistently outperforms both prompt engineering and SFT in both $\Delta$MRR and $\Delta$DIR, achieving the highest $\Delta$MRR value of 0.0073 and highest $\Delta$DIR value of 2.7944, both of which are statistically significant under a 95\% confidence interval. SFT achieves the second best $\Delta$DIR score of 1.6382 but does have a low value for $\Delta$MRR of 0.0022. 

For few-shot prompting, PPO maintains a strong $\Delta$DIR of 1.9267 which is relatively similar to prompt engineering's $\Delta$DIR value of 1.9133, but does exhibit a negative $\Delta$MRR value. SFT performs worst here, with both lower $\Delta$MRR and $\Delta$DIR values. 

\subsection{Ablation Studies}
\subsubsection{Varying K}
We conduct ablation studies by varying the value of k to examine how the amount of ads retrieved affects the metrics $\Delta$MRR@k and $\Delta$DIR@k across models. 

\begin{table}[H]
\centering
\begin{tabular}{lcc}
\toprule
\textbf{$k$} & \textbf{$\Delta$MRR@k} & \textbf{$\Delta$DIR@k} \\
\midrule
1 & 0.00 & 0.6293 \\
3 & 0.00 & -0.558 \\
5 & 0.00 & -0.0807 \\
10 & 0.00 & -0.2844 \\
20 & 0.00 & 0.2288 \\
30 & 0.00 & 0.1246 \\
\bottomrule
\end{tabular}
\caption{$\Delta$MRR and $\Delta$DIR values across varying k values for the zero-rewrite baseline}
\label{tab:mrr-dir-zero}
\end{table}

The $\Delta$MRR for the zero-rewrite control is always 0.00 since the ad content remains unchanged and therefore has identical similarity to the query. The $\Delta$DIR, while expected to be near zero, may have slight fluctuations as the nondeterministic nature of LLM generated responses can include different ads across runs even when the input remains the same. 

\onecolumn
\begin{table}[H]
    \centering
    \begin{minipage}[t]{0.48\linewidth}
        \centering
        \begin{tabular}{cccc}
        \toprule
        \textbf{$k$} & \textbf{Prompt Eng.} & \textbf{SFT} & \textbf{PPO} \\
        \midrule
        1 & 0.0033 & 0.0035 & 0.0072 \\
        3 & 0.0051 & 0.0026 & 0.0072 \\
        5 & 0.0051 & 0.0022 & 0.0072 \\
        10 & 0.0055 & 0.0036 & 0.0085 \\
        20 & 0.0065 & 0.0038 & 0.0082 \\
        30 & 0.0064 & 0.0036 & 0.0082 \\
        \bottomrule
        \end{tabular}
        \caption{$\Delta$MRR values for different $k$ across models with instruction-based prompting}
        \label{tab:mrr-k-valuesinst}
    \end{minipage}
    \hfill
    \begin{minipage}[t]{0.48\linewidth}
        \centering
        \begin{tabular}{cccc}
        \toprule
        \textbf{$k$} & \textbf{Prompt Eng.} & \textbf{SFT} & \textbf{PPO} \\
        \midrule
        1 & 0.0062 & -0.0003 & -0.0000 \\
        3 & 0.0068 & -0.0017 & -0.0006 \\
        5 & 0.0067 & -0.0017 & -0.0008 \\
        10 & 0.0081 & -0.0015 & -0.0005 \\
        20 & 0.0088 & -0.0014 & -0.0008 \\
        30 & 0.0090 & -0.0017 & -0.0010 \\
        \bottomrule
        \end{tabular}
        \caption{$\Delta$MRR values for different $k$ across models with few-shot prompting}
        \label{tab:mrr-k-valuesfew}
    \end{minipage}
\end{table}

\vspace{-1em}

\begin{table}[H]
    \centering
    \begin{minipage}[t]{0.48\linewidth}
        \centering
        \begin{tabular}{cccc}
        \toprule
        \textbf{$k$} & \textbf{Prompt Eng.} & \textbf{SFT} & \textbf{PPO} \\
        \midrule
        1 & 0.4619 & 0.7773 & 1.8217 \\
        3 & 0.8898 & 1.3500 & 2.2337 \\
        5 & 0.9061 & 1.6382 & 2.7944 \\
        10 & 0.4598 & 1.0754 & 2.2342 \\
        20 & -0.8517 & -0.7896 & -0.5477 \\
        30 & -0.4629 & -0.7658 & -0.5215 \\
        \bottomrule
        \end{tabular}
        \caption{$\Delta$DIR values for different $k$ across models with instruction-based prompting}
        \label{tab:dir-k-valuesinst}
    \end{minipage}
    \hfill
    \begin{minipage}[t]{0.48\linewidth}
        \centering
        \begin{tabular}{cccc}
        \toprule
        \textbf{$k$} & \textbf{Prompt Eng.} & \textbf{SFT} & \textbf{PPO} \\
        \midrule
        1 & 0.0139 & 0.8113 & 1.5303 \\
        3 & 0.6524 & 0.8892 & 2.0755 \\
        5 & 1.9133 & 1.4199 & 1.9267 \\
        10 & 0.5740 & 0.9773 & 1.4795 \\
        20 & -0.9622 & -0.9881 & -0.9081 \\
        30 & -0.4483 & -0.4321 & -0.6708 \\
        \bottomrule
        \end{tabular}
        \caption{$\Delta$DIR values for different $k$ across models with few-shot prompting}
        \label{tab:dir-k-valuesfew}
    \end{minipage}
\end{table}

\vspace{-0.5em}

\begin{figure}[H]
    \centering
    \begin{minipage}[t]{0.48\linewidth}
        \centering
        \includegraphics[width=\linewidth]{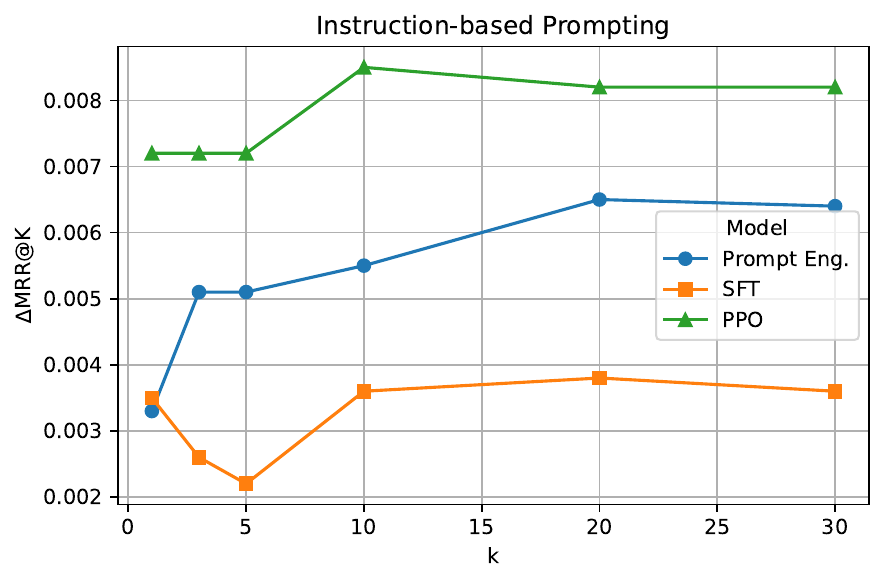}
        \vspace{-2em}
        \caption{$\Delta$MRR metric for instruction-based prompting across $k$ values.}
    \end{minipage}
    \hfill
    \centering
    \begin{minipage}[t]{0.48\linewidth}
        \centering
        \includegraphics[width=\linewidth]{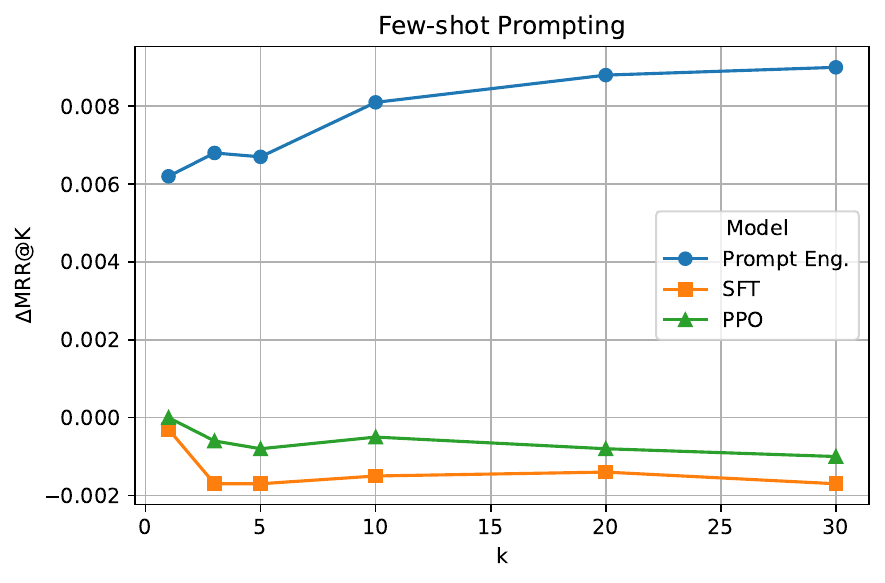}
        \vspace{-2em}
        \caption{$\Delta$MRR metric for few-shot prompting across $k$ values.}
    \end{minipage}
\end{figure}
\vspace{-1em}
\begin{figure}[H]
    \begin{minipage}[t]{0.48\linewidth}
        \centering
        \includegraphics[width=\linewidth]{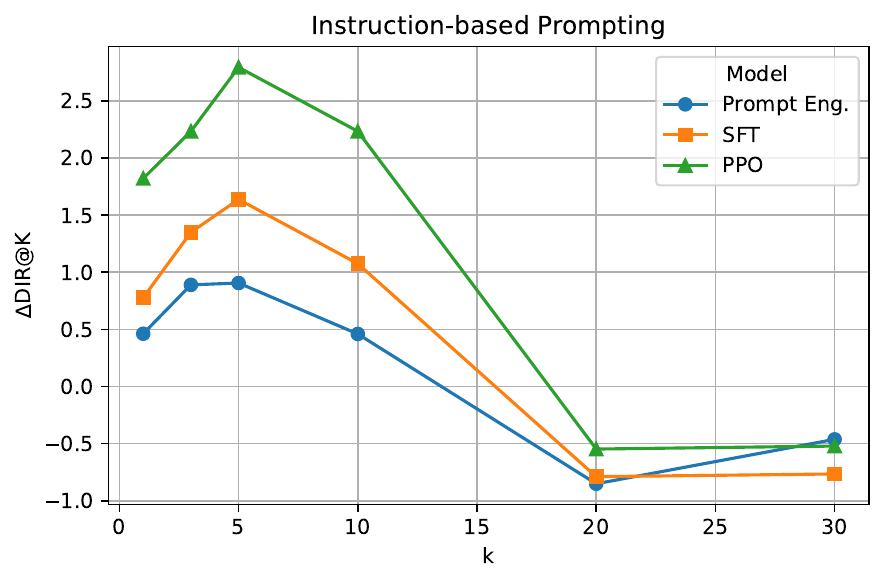}
        \vspace{-2em}
        \caption{$\Delta$DIR metric for instruction-based prompting across $k$ values.}
    \end{minipage}
    \hfill
    \begin{minipage}[t]{0.48\linewidth}
        \centering
        \includegraphics[width=\linewidth]{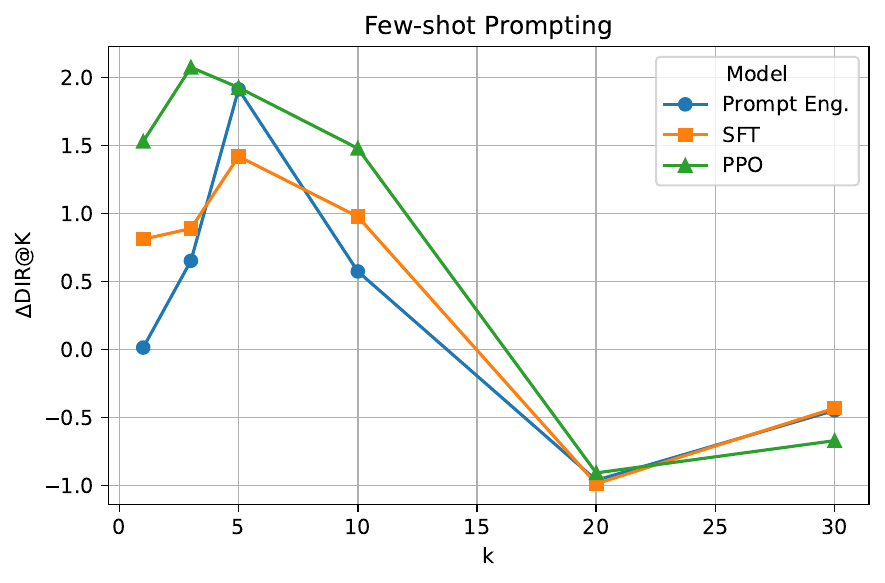}
        \vspace{-2em}
        \caption{$\Delta$DIR metric for few-shot prompting across $k$ values.}
    \end{minipage}
\end{figure}
\FloatBarrier

\begin{multicols}{2}

\paragraph{$\Delta$MRR Trends (Tables \ref{tab:mrr-k-valuesinst}, \ref{tab:mrr-k-valuesfew}):}
\begin{itemize}
\vspace{-1em}
\item For instruction-based prompting, PPO maintains consistent performance across all values of k, with $\Delta$MRR peaking at k = 10 (0.0085), surpassing the values of prompt engineering and SFT. 
\item For few-shot prompting, only prompt engineering has a positive value of $\Delta$MRR. Both SFT and PPO have negative or near zero $\Delta$MRR values suggesting that few-shot rewriting may decrease the ad's relevance to the query. 
\end{itemize}

\paragraph{$\Delta$DIR Trends (Tables \ref{tab:dir-k-valuesinst}, \ref{tab:dir-k-valuesfew}):}
\begin{itemize}
\vspace{-1em}
\item Under instruction-based prompting, PPO has the highest values of $\Delta$DIR, peaking at k = 5 (2.7944), surpassing both prompt engineering and SFT. Beyond k = 20, all three models exhibited negative values of DIR. 
\item For few-shot prompting, $\Delta$DIR is the highest for PPO at k = 3 (2.0755), but becomes negative beyond k = 20. 
\end{itemize}

\subsubsection{Loss Function}

To understand the impact of each component in our loss function, we conduct an ablation study across six different weightings of the composite loss used during PPO for instruction-based and few-shot prompting: 

$L_{\text{total}} = \alpha L_{\text{rel\_gain}} + \beta L_{\text{triplet}} + \gamma L_{\text{fidelity}}$

The weightings tested are:
\vspace{-1em}

\begin{itemize}
    \item \textbf{Equal weighting} ($1{:}1{:}1$) – baseline
    \item $35{:}45{:}20$ – emphasizing triplet loss
    \item $45{:}35{:}20$ – emphasizing relevance gain
    \item $45{:}20{:}35$ – emphasizing content fidelity
    \item $20{:}45{:}35$ – balanced triplet and fidelity
    \item $30{:}60{:}10$ – heavily emphasizing triplet loss
\end{itemize}
\end{multicols}

\begin{table}[H]
    \centering
    \begin{minipage}[t]{0.48\linewidth}
        \centering
        \begin{tabular}{ccccc}
        \toprule
        \textbf{$k$} & \textbf{$\Delta$MRR in.} & \textbf{$\Delta$DIR in.}  & \textbf{$\Delta$MRR fe.} & \textbf{$\Delta$DIR fe.}\\
        \midrule
        1 & 0.0072 & 1.8217 & 0.0000 & 1.5303 \\
        3 & 0.0072 & 2.2337 & -0.0006 & 2.0755 \\
        5 & 0.0072 & 2.7944 & -0.0008 & 1.9267 \\
        10 & 0.0085 & 2.2342 & -0.0005 & 1.4795 \\
        20 & 0.0082 & -0.5477 & -0.0008 & -0.9081 \\
        30 & 0.0082 & -0.5215 & -0.0010 & -0.6708 \\
        \bottomrule
        \end{tabular}
        \caption{$\Delta$MRR@k and $\Delta$DIR@k values at $\alpha=1$, $\beta=1$, and $\gamma=1$}
        \label{tab:ppo1:1:1}
    \end{minipage}
    \hfill
    \begin{minipage}[t]{0.48\linewidth}
        \centering
        \begin{tabular}{ccccc}
        \toprule
        \textbf{$k$} & \textbf{$\Delta$MRR in.} & \textbf{$\Delta$DIR in.}  & \textbf{$\Delta$MRR fe.} & \textbf{$\Delta$DIR fe.}\\
        \midrule
        1 & -0.0061 & 0.7782 & -0.0031 & 1.531 \\
        3 & 0.0070 & 2.6131 & -0.0053 & 1.3385 \\
        5 & 0.0058 & 2.7644 & -0.0053 & 2.5133 \\
        10 & 0.0058 & 2.3594 & -0.0058 & 3.0537 \\
        20 & 0.0061 & 0.3193 & -0.0058 & 0.6826 \\
        30 & 0.0050 & 0.5871 & -0.0061 & 0.7782 \\
        \bottomrule
        \end{tabular}
        \caption{$\Delta$MRR@k and $\Delta$DIR@k values at $\alpha=0.35$, $\beta=0.45$, and $\gamma=0.20$}
        \label{tab:ppo354520}
    \end{minipage}
\end{table}

\vspace{-1em}

\begin{table}[H]
    \centering
    \begin{minipage}[t]{0.48\linewidth}
        \centering
        \begin{tabular}{ccccc}
        \toprule
        \textbf{$k$} & \textbf{$\Delta$MRR in.} & \textbf{$\Delta$DIR in.}  & \textbf{$\Delta$MRR fe.} & \textbf{$\Delta$DIR fe.}\\
        \midrule
        1 & -0.0051 & 2.0068 & -0.0045 & 2.5359 \\
        3 & -0.0048 & 2.3990 & -0.0074 & 2.1676 \\
        5 & -0.0047 & 2.9368 & -0.0087 & 2.1994 \\
        10 & -0.0037 & 2.5265 & -0.0090 & 2.0288 \\
        20 & -0.0033 & 0.6943 & -0.0089 & 0.9471 \\
        30 & -0.0033 & 0.7806 & -0.009 & 0.7327 \\
        \bottomrule
        \end{tabular}
        \caption{$\Delta$MRR@k and $\Delta$DIR@k values at $\alpha=0.45$, $\beta=0.35$, and $\gamma=0.20$}
        \label{tab:ppo453520}
    \end{minipage}
    \hfill
    \begin{minipage}[t]{0.48\linewidth}
        \centering
        \begin{tabular}{ccccc}
        \toprule
        \textbf{$k$} & \textbf{$\Delta$MRR in.} & \textbf{$\Delta$DIR in.}  & \textbf{$\Delta$MRR fe.} & \textbf{$\Delta$DIR fe.}\\
        \midrule
        1 & 0.0046 & 2.2036 & 0.0002 & 1.7101 \\
        3 & 0.0042 & 2.3052 & -0.0021 & 1.8560 \\
        5 & 0.0045 & 2.5554 & -0.0034 & 2.5287 \\
        10 & 0.0056 & 2.2332 & -0.0043 & 2.177 \\
        20 & 0.0059 & 0.9995 & -0.0043 & 0.9982 \\
        30 & 0.0058 & 0.9697 & -0.0045 & 0.7180 \\
        \bottomrule
        \end{tabular}
        \caption{$\Delta$MRR@k and $\Delta$DIR@k values at $\alpha=0.45$, $\beta=0.20$, and $\gamma=0.35$}
        \label{tab:ppo452035}
    \end{minipage}
\end{table}

\vspace{-1em}

\begin{table}[H]
    \centering
    \begin{minipage}[t]{0.48\linewidth}
        \centering
        \begin{tabular}{ccccc}
        \toprule
        \textbf{$k$} & \textbf{$\Delta$MRR in.} & \textbf{$\Delta$DIR in.}  & \textbf{$\Delta$MRR fe.} & \textbf{$\Delta$DIR fe.}\\
        \midrule
        1 & 0.0028 & 1.1920 & -0.0014 & 2.2464 \\
        3 & 0.0031 & 1.2890 & -0.0025 & 2.4326 \\
        5 & 0.0036 & 2.6791 & -0.0040 & 2.8285 \\
        10 & 0.0036 & 2.6109 & -0.0048 & 2.0644 \\
        20 & 0.0039 & 1.0337 & -0.0050 & 1.5736 \\
        30 & 0.0038 & 0.9374 & -0.0053 & 0.9407 \\
        \bottomrule
        \end{tabular}
        \caption{$\Delta$MRR@k and $\Delta$DIR@k values at $\alpha=0.20$, $\beta=0.45$, and $\gamma=0.35$}
        \label{tab:ppo4204535}
    \end{minipage}
    \hfill
    \begin{minipage}[t]{0.48\linewidth}
        \centering
        \begin{tabular}{ccccc}
        \toprule
        \textbf{$k$} & \textbf{$\Delta$MRR in.} & \textbf{$\Delta$DIR in.}  & \textbf{$\Delta$MRR fe.} & \textbf{$\Delta$DIR fe.}\\
        \midrule
        1 & 0.0077 & 1.7394 & -0.0015 & 2.0998 \\
        3 & 0.0078 & 2.4325 & -0.0023 & 2.1454 \\
        5 & 0.0066 & 2.6691 & -0.0044 & 2.9004 \\
        10 & 0.0068 & 2.2473 & -0.0045 & 2.2760 \\
        20 & 0.0068 & 0.8375 & -0.0046 & 1.7480 \\
        30 & 0.0068 & 0.6761 & -0.0046 & 1.7158 \\
        \bottomrule
        \end{tabular}
        \caption{$\Delta$MRR@k and $\Delta$DIR@k values at $\alpha=0.30$, $\beta=0.60$, and $\gamma=0.10$}
        \label{tab:ppo306010}
    \end{minipage}
\end{table}

\vspace{-0.5em}

\begin{figure}[H]
    \centering
    \begin{minipage}[t]{0.48\linewidth}
        \centering
        \includegraphics[width=\linewidth]{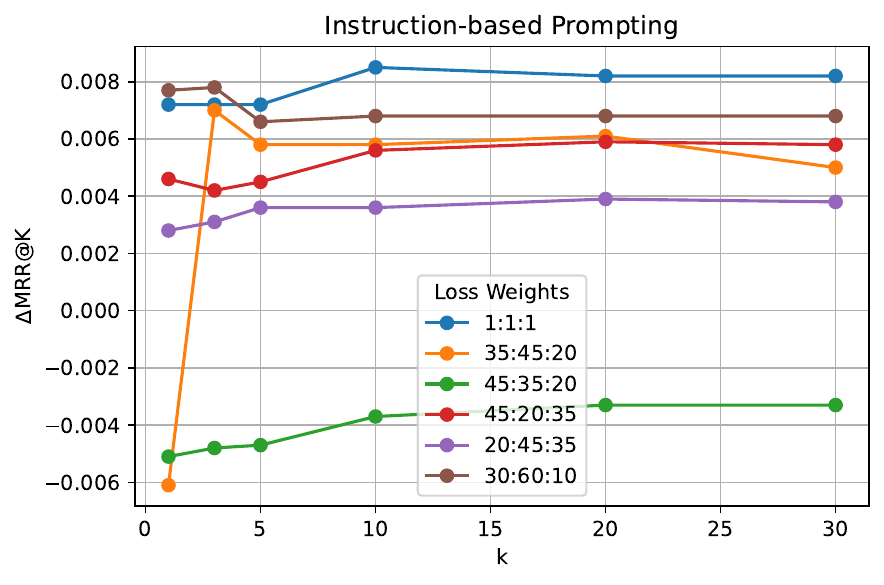}
        \vspace{-2em}
        \caption{$\Delta$MRR@k metric for instruction-based prompting across varying loss weights.}
    \end{minipage}
    \hfill
    \centering
    \begin{minipage}[t]{0.48\linewidth}
        \centering
        \includegraphics[width=\linewidth]{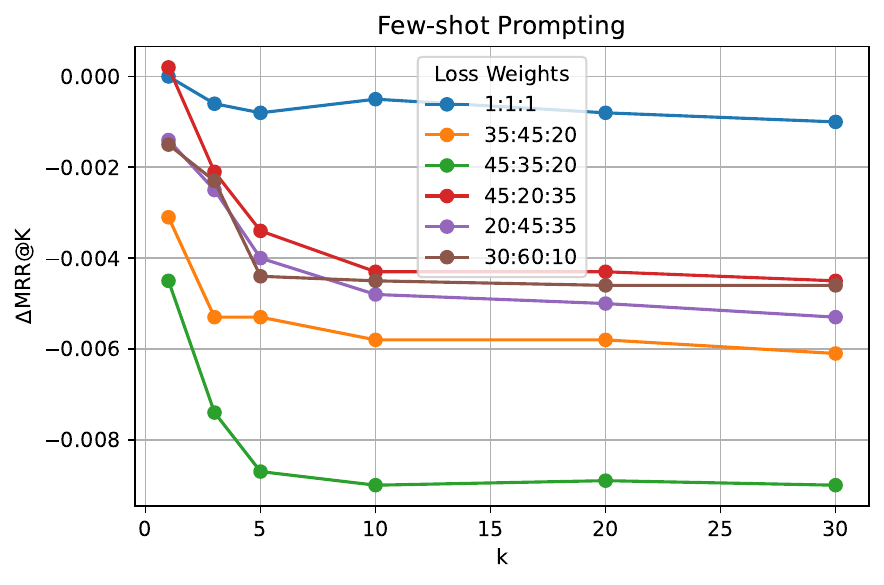}
        \vspace{-2em}
        \caption{$\Delta$MRR@k metric for few-shot prompting across varying loss weights.}
    \end{minipage}
\end{figure}
\vspace{-2em}
\begin{figure}[H]
    \begin{minipage}[t]{0.48\linewidth}
        \centering
        \includegraphics[width=\linewidth]{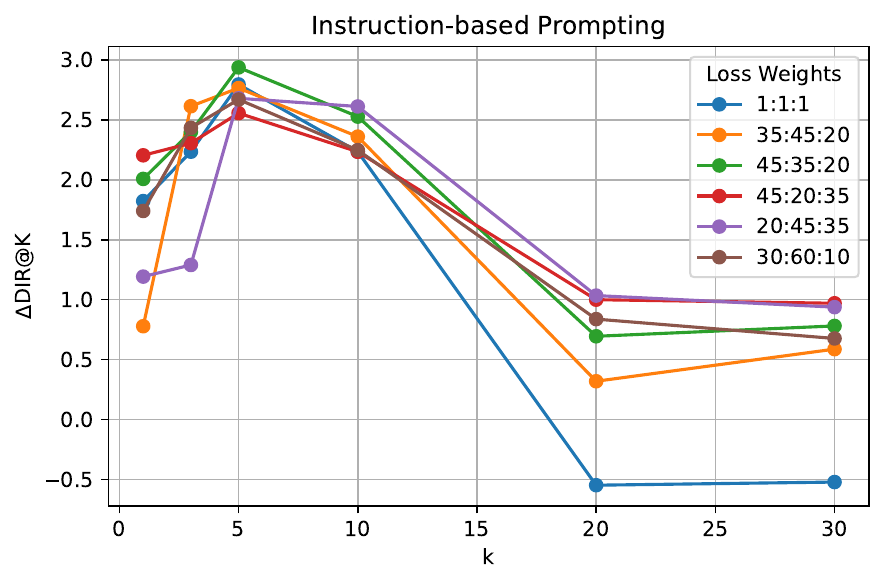}
        \vspace{-2em}
        \caption{$\Delta$DIR@k metric for instruction-based prompting across varying loss weights..}
    \end{minipage}
    \hfill
    \begin{minipage}[t]{0.48\linewidth}
        \centering
        \includegraphics[width=\linewidth]{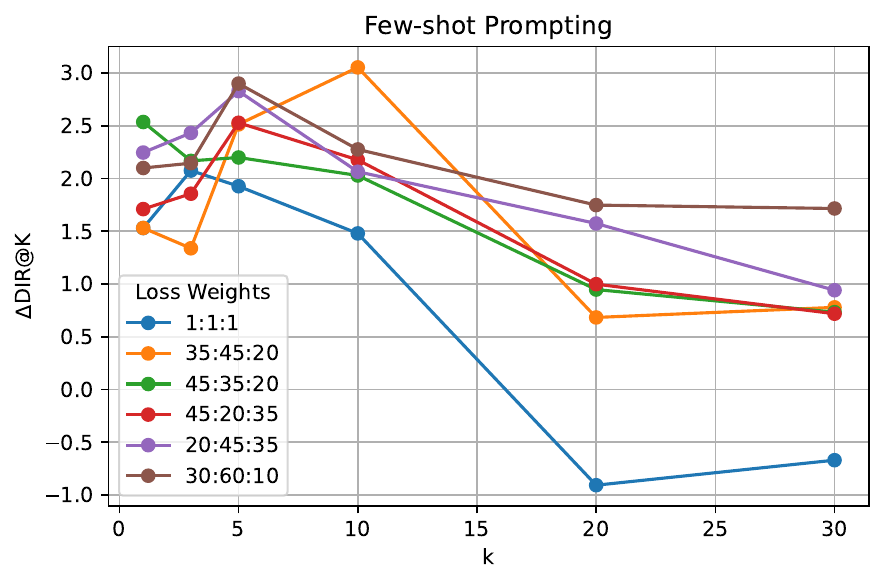}
        \vspace{-2em}
        \caption{$\Delta$MRR@k metric for few-shot prompting across varying loss weights.}
    \end{minipage}
\end{figure}
\FloatBarrier

\begin{multicols}{2}

\paragraph{$\Delta$MRR Trends (~\cref{tab:ppo1:1:1,tab:ppo354520,tab:ppo453520,tab:ppo452035,tab:ppo4204535,tab:ppo306010} ):}
\begin{itemize}
    \vspace{-1em}
    \item For \textit{instruction-based} prompting, the equal weighting ($1{:}1{:}1$) achieves the most consistent positive $\Delta\text{MRR}@5$ performance ($0.0072$), outperforming most alternative weightings. The $45{:}35{:}20$ weighting shows negative $\Delta\text{MRR}$ values ($-0.0047$ at $k{=}5$) despite achieving the highest $\Delta\text{DIR}$ performance.
    \item For \textit{few-shot} prompting, most weightings result in negative $\Delta\text{MRR}$ values, with the equal weighting ($1{:}1{:}1$) maintaining the least negative impact ($\Delta\text{MRR}@5=-0.0008$). This suggests that few-shot prompting may inherently trade ranking performance for inclusion improvements.
\end{itemize}

\paragraph{$\Delta$DIR Trends (~\cref{tab:ppo1:1:1,tab:ppo354520,tab:ppo453520,tab:ppo452035,tab:ppo4204535,tab:ppo306010} ):}
\begin{itemize}
    \vspace{-1em}
    \item Under \textit{instruction-based} prompting, the $45{:}35{:}20$ weighting achieves peak $\Delta\text{DIR}@5$ performance ($2.9368$), surpassing the equal-weighting baseline ($2.7944$). The $35{:}45{:}20$ weighting also shows strong inclusion performance ($2.7644$ at $k{=}5$).
    \item For \textit{few-shot} prompting, the $30{:}60{:}10$ weighting performs well on inclusion metrics, achieving $\Delta\text{DIR}@5=2.9004$ and the $35{:}45{:}20$ weighting achieves strong performance at $k{=}10$ ($\Delta\text{DIR}@10 = 3.0537$).
\end{itemize}

\section{Discussion}
\paragraph{Main comparison at $k{=}5$}
Our main claim is that \textit{PPO simultaneously boosts rank and inclusion under instruction prompting}.  
Table~\ref{tab:mrr-dir-valuesinst} confirms this: \textbf{PPO lifts} $\Delta$MRR@5 from $0.0051$ (prompt engineering) to $0.0073$ and increases $\Delta$DIR@5 from $0.9061$ to $2.7944$. 
Supervised fine-tuning reduces rank to $0.0022$ yet still improves inclusion to $1.6382$, indicating that content-fidelity supervision alone cannot fully overcome retriever bias but that lower MRR does not correlate with higher inclusion. 

For the \emph{few-shot} prompting (Table~\ref{tab:mrr-dir-valuesfew}), PPO continues to increase $\Delta$DIR (+1.93) but lowers $\Delta$MRR (–$0.0008$), suggesting that rewriting improves the chance of an ad being integrated by an LLM but not its relevance to a query. The rank dip persists for all $k$ values in Table~\ref{tab:mrr-k-valuesfew}, whereas $\Delta$DIR remains positive until $k{=}20$, matching typical RAG first-page windows \citep{Lewis20}.

\paragraph{Sensitivity to cut-off $k$}
\vspace{-0.5em}
Under instruction prompting (Tables~\ref{tab:mrr-k-valuesinst}, \ref{tab:dir-k-valuesinst}), the metrics for PPO usually peak at around k = 5 to k = 10, decreasing beyond k = 20. This underscores that rewriting may become less effective in retrieval ranking and inclusion as less relevant ads are being ranked and the LLM has more ads to choose from. 

\paragraph{Ablation: $k$ value}
\vspace{-0.5em}
Optimal retrieval occurs at moderate values of k = 5 to k = 10 because large values of k introduce noise and diminish relevance. Removing PPO and retaining only SFT drops $\Delta$MRR by $60$–$70\%$ yet still improves $\Delta$DIR, confirming that \textit{content-fidelity supervision alone} cannot reliably raise rank \citep{Dong24}. 

\paragraph{Ablation: Loss weighting}
\vspace{-0.5em}
The equal weighting ($1{:}1{:}1$) provides the most balanced performance across both ranking and inclusion metrics, suggesting that all three components contribute meaningfully to effective ad rewriting \citep{Golnari2023Adaptive,Chen2018GradNorm}. $L_{\text{rel\_gain}}$ directly optimizes query–document similarity, explaining why higher $\alpha$ values can improve inclusion rates \citep{Hou2021AdaptiveTriplet}. $L_{\text{triplet}}$ helps ads outperform competing documents, which is crucial in competitive retrieval scenarios, as shown by the high inclusion (few-shot) and relatively high $\Delta$MRR (instruction-based) in Tables \ref{tab:ppo354520} and \ref{tab:ppo306010}. $L_{\text{fidelity}}$ prevents semantic drift, but can limit the model’s ability to make strategic keyword choices \citep{Chaturvedi2022Faithfulness,Li2024WalkTheTalk}.

\paragraph{Why does PPO help?}
\vspace{-0.5em}
PPO optimizes a composite reward balancing query similarity, distractor margin, and fidelity, leveraging its stability for LLM alignment \citep{Schulman17,Ram23}. This surfaces high-IDF keywords valued by the retriever while preserving semantics, hence the joint lift in $\Delta$MRR and $\Delta$DIR \citep{Ziegler20}. 

\paragraph{Inclusion vs.\ ranking.}
\vspace{-0.5em}
An important consideration for practical deployment is that inclusion in LLM responses ($\Delta\text{DIR}$) may be more commercially valuable than ranking position ($\Delta\text{MRR}$).  
Recent RAG-evaluation studies emphasise that the text actually surfaced in the answer, rather than its raw retrieval rank, drives user value and system quality \citep{Yu2024RAGSurvey,Madaan2024LLMEval}.  
Because users never see the internal ranking scores but do see which passages are quoted, optimizing $\Delta\text{DIR}$ is often more aligned with real-world objectives \citep{Es2023RAGAS}.  
Our own results echo this: configurations such as $45{:}35{:}20$ (instruction) and $35{:}45{:}20$ (few-shot) push inclusion strongly even when ranking gains are modest or negative.

\paragraph{Strategy-specific}
\vspace{-0.5em}
Instruction prompting typically supplies richer task signals than few-shot demonstrations, which explains why our instruction-based run with $\alpha{=}0.45,\beta{=}0.35,\gamma{=}0.20$ achieves the highest $\Delta\text{DIR}@5$ \citep{Wei2022CoT,Wang2022CoTStudy}.  
In contrast, few-shot exemplars anchor the model to brevity and marketing tone; the fidelity weight over-penalizes lexical deviation, reducing keyword coverage, and hurting rank, a pattern also observed in ranking work on few-shot prompting \citep{Sinhababu24} and lexical overlap analyses of dense retrievers \citep{Ram23}. Few-shot setups benefit from emphasizing hard-negative discrimination ($\beta$) as in the $35{:}45{:}20$ setting, mirroring findings that chain-of-thought exemplars can over-anchor lexical style and hurt precision unless balanced by discriminative losses \citep{Zhang2023PlanSolve}.

\begin{itemize}
    \vspace{-1.15em}
    \item \textbf{Instruction-based prompting:} $\alpha=0.45, \beta=0.35, \gamma=0.20$ achieves peak $\Delta\text{DIR}@5 = 2.9368$.
    \vspace{-1.75em}
    \item \textbf{Few-shot prompting:} $\alpha=0.35,\ \beta=0.45,\ \gamma=0.20$ achieves peak $\Delta\text{DIR}@10 = 3.0537$.
\end{itemize}
\vspace{-0.5em}
The results demonstrate a clear \textbf{trade-off between ranking and inclusion performance}. Most alternative weightings improve inclusion at the expense of ranking performance, with only the equal weighting maintaining positive ranking improvements while achieving competitive inclusion rates. Practitioners should therefore pick weights according to business goals: equal weighting for balanced improvements, or inclusion-focused weightings when ad visibility in LLM responses is the primary concern.

\section{Conclusion}
In this study, we introduced a novel approach to improve ad visibility in LLM retrieval systems through content rewriting, without modifying the retriever itself. Our framework leverages Proximal Policy Optimization to improve retrieval ranking and inclusion in the LLM response. By creating a custom composite loss that maximizes semantic relevance and content fidelity, we demonstrate how language models can be fine-tuned to make commercially effective rewrites. Empirically, we show that our PPO enhanced models generally outperform both supervised fine-tuning and prompt engineering baselines on two key metrics, $\Delta$MRR@k and $\Delta$DIR@k, over multiple k values and two prompting strategies, suggesting that LLM rewriting and optimization is an effective approach to improve ad visibility in language model pipelines. As LLMs continue to power search, recommendation, and generation pipelines, our results indicate that phrasing optimization can improve retrieval output for black-box systems. Our hope is that this method can serve as a foundation for future work on retrieval-aware and LLM alignment, especially in domains where visibility directly influences utility or monetization.


\bibliography{citations}
\bibliographystyle{icml2025}

\newpage
\end{multicols}{}
\appendix
\section{Appendix}
\subsection{Prompt Design}
\begin{tcolorbox}[title=Black-Box LLM that Integrates the Ads Prompt,
                  colback=gray!5,
                  colframe=gray!60,
                  breakable,
                  listing only,
                  listing engine=listings, 
                  listing options={
                    basicstyle=\ttfamily\small,
                    breaklines=true,       
                    columns=fullflexible   
                  }]

You are a helpful and knowledgeable assistant. Below is a list of product advertisements. Your task is to respond to the user's query in a natural and informative way, promoting one or two relevant products. Include brand and url into your response without revealing that you have access to product ads. 
\vspace{1em}

At the end of your response (not visible to the user), list which documents you included clearly only using the format: 'id: ..., id: ..., ...'.
\vspace{1em}

USER QUERY: {query}

RETRIEVED COMMERCIAL ADS:
{context}
\vspace{1em}

Please provide a helpful, informative response directed to the user based on the above information.
\end{tcolorbox}

\begin{tcolorbox}[title=General prompting strategy,
                  colback=gray!5,
                  colframe=gray!60,
                  breakable,
                  listing only,
                  listing engine=listings, 
                  listing options={
                    basicstyle=\ttfamily\small,
                    breaklines=true,       
                    columns=fullflexible   
                  }]

You are given an advertisement. Your task is to rewrite it so that its ranking in retrieval and inclusion in LLM response improves. Focus on semantic relevance and matching the user’s likely search intent. 
\vspace{1em}

Original Ad: {ad}
\vspace{1em}

Respond with the improved version:
\vspace{1em}

Title: ...
\vspace{1em}

Description: ...

\end{tcolorbox}

\begin{tcolorbox}[title=Instruction-based Prompting with CoT (Zero-Shot),
                  colback=gray!5,
                  colframe=gray!60,
                  breakable,
                  listing only,
                  listing engine=listings, 
                  listing options={
                    basicstyle=\ttfamily\small,
                    breaklines=true,       
                    columns=fullflexible   
                  }]

You are given an advertisement. Your task is to rewrite the ad so that its ranking in retrieval and inclusion in LLM responses improves. Focus on semantic relevance and matching the user's likely search intent.
\vspace{1em}

Original Ad: {ad}
\vspace{1em}

Think step by step first, then provide the improved version.
\vspace{1em}

Respond with the improved version at the end of your response **only** in the following format:
\vspace{1em}

Thinking: ...
\vspace{0.5em}

Title: ...
\vspace{0.35em}

Description: ...

\end{tcolorbox}

\begin{tcolorbox}[title=Few-Shot Prompting with CoT,
                  colback=gray!5,
                  colframe=gray!60,
                  breakable,
                  listing only,
                  listing engine=listings, 
                  listing options={
                    basicstyle=\ttfamily\small,
                    breaklines=true,      
                    columns=fullflexible   
                  }]

Rewrite the advertisement so that it ranks better in retrieval and its inclusion in LLM responses improves. Here are two examples:
\vspace{1em}

Example 1
\vspace{0.5em}

Original Ad: Title: Yoga Pants Description: Affordable yoga pants for women, available in multiple colors.
\vspace{0.5em}

Reasoning: The phrase "affordable yoga pants" is generic. Adding activity-specific and quality-based terms may help.
\vspace{0.5em}

Rewritten Ad: Title: High-performance women’s yoga leggings Description: great yoga pants for training and Pilates – breathable, colorful, and comfortable.
\vspace{1em}

Example 2
\vspace{0.5em}

Original Ad: Title: mugs Description: Buy custom mugs with your name.
\vspace{0.5em}

Reasoning: This lacks variety and emotional appeal. Including gifting context and materials can help retrieval.
\vspace{0.5em}

Rewritten Ad: Title: Personalized ceramic mugs Description: perfect gifts with names, photos, or messages.
\vspace{1em}

Your turn
\vspace{0.5em}

Original Ad: {ad}
\vspace{0.5em}

Reasoning: 
\vspace{0.5em}

Rewritten Ad: 

\end{tcolorbox}

\subsection{PPO Hyper-Parameters}
Table~\ref{tab:ppo-hp} lists every setting required to reproduce our RL stage.

\begin{table}[h!]
\centering
\small
\begin{tabular}{lcc}
\toprule
\textbf{Category} & \textbf{Symbol / Key} & \textbf{Value} \\
\midrule
Clip range & $\epsilon$ & 0.2 \\
KL penalty & $\beta$ & 0.1 \\
PPO epochs / batch & — & 4 \\
Reward discount & $\gamma$ & 1.0 \\
\midrule
Micro-batch size & $m$ & 4 \\
Grad-accumulation & $g$ & 4 \\
\textit{Effective} batch & $m\times g$ & 16 \\
Learning rate & $\eta$ & $1\times10^{-5}$ \\
Scheduler & — & Cosine \\
Warm-up ratio & — & 0.10 \\
\midrule
Train epochs & — & 3 \\
Dataset size & $N$ & 10\,000 ads \\
\textbf{Total PPO updates}$^{\dagger}$ & $\lceil 3N / (16\!\times\!N_{\text{GPU}}) \rceil$ & 1 875 (1 GPU) \\
\bottomrule
\end{tabular}
\vspace{0.4em}
\caption{PPO settings for LLaMA-3.1-8B. $^{\dagger}$Divide by $N_{\text{GPU}}$ to scale.}
\label{tab:ppo-hp}
\end{table}

\subsection{Code and Data Availability}

To foster reproducibility while preserving anonymity, we release our
implementation in two static repositories:

\begin{itemize}
  \item \url{https://anonymous.4open.science/r/ad-ppo-lora-2706}
  \item \url{https://anonymous.4open.science/r/ad-doc-reranker-57C6}
\end{itemize}

\end{document}